\documentclass[letterpaper]{article} 
\usepackage{aaai24}  
\usepackage{times}  
\usepackage{helvet}  
\usepackage{courier}  
\usepackage[hyphens]{url}  
\usepackage{graphicx} 
\usepackage{natbib}  
\usepackage{caption} 
\usepackage{algorithm}
\usepackage{upgreek}
\usepackage{algorithmic}

\usepackage{newfloat}
\usepackage{listings}
\DeclareCaptionStyle{ruled}{labelfont=normalfont,labelsep=colon,strut=off} 
\floatstyle{ruled}
\newfloat{listing}{tb}{lst}{}
\floatname{listing}{Listing}
\usepackage{multirow}
\usepackage{booktabs}
\usepackage{amssymb}
\usepackage{amsmath}
\usepackage{subfigure}

\title{Fine-Grained Graph Representation Learning for Heterogeneous Mobile Networks with Attentive Fusion and Contrastive Learning}
\author{
Shengheng~Liu\textsuperscript{\rm 1, 2 *},
Tianqi~Zhang\textsuperscript{\rm 1},
Ningning~Fu\textsuperscript{\rm 1},
and Yongming~Huang\textsuperscript{\rm 1, 2 *}
}
\affiliations{
    \textsuperscript{\rm 1}National Mobile Communications Research Laboratory, Southeast University, Nanjing, China\\
    \textsuperscript{\rm 2}Purple Mountain Laboratories, Nanjing, China\\

  \{s.liu; huangym\}@seu.edu.cn
}

\usepackage{bibentry}

\begin{document}

\maketitle

\begin{abstract}
AI becomes increasingly vital for telecom industry, as the burgeoning complexity of upcoming mobile communication networks places immense pressure on network operators. While there is a growing consensus that intelligent network self-driving holds the key, it heavily relies on expert experience and knowledge extracted from network data. In an effort to facilitate convenient analytics and utilization of wireless big data, we introduce the concept of knowledge graphs into the field of mobile networks, giving rise to what we term as wireless data knowledge graphs (WDKGs). However, the heterogeneous and dynamic nature of communication networks renders manual WDKG construction both prohibitively costly and error-prone, presenting a fundamental challenge. In this context, we propose an unsupervised \textbf{d}ata-and-\textbf{m}odel driven \textbf{g}raph \textbf{s}tructure \textbf{l}earning (DMGSL) framework, aimed at automating WDKG refinement and updating. Tackling WDKG heterogeneity involves stratifying the network into homogeneous layers and refining it at a finer granularity. Furthermore, to capture WDKG dynamics effectively, we segment the network into static snapshots based on the coherence time and harness the power of recurrent neural networks to incorporate historical information. Extensive experiments conducted on the established WDKG demonstrate the superiority of the DMGSL over the baselines, particularly in terms of node classification accuracy.
\end{abstract}

\section{Introduction}

The emerging trend of convergence between AI and wireless technology is expected to bring along new research opportunities and better connectivity for people. Now commercial usage of 5G has reached maturity in the leading markets and has sparked a growing appetite for new services that imply extremely stringent requirements. The rapid evolution of networks’ capabilities has introduced significant structural complexity, posing challenges for network management and maintenance. As such, the spotlight is on network automation as a prominent trend for forthcoming 6G networks projected for launch by 2030 \cite{You2023}, which integrates essential functions such as self-configuring, self-optimizing, self-protecting and self-healing \cite{Chi2023}. Achieving such a high degree of network automation requires a fusion of knowledge from both physical models and network big data. Fortunately, knowledge graphs (KGs), a powerful tool to integrate knowledge and data, offer a promising solution for network automation. 

While efforts have been made to establish wireless knowledge graphs (WDKGs) \cite{Huang2024}, the existing ones are constructed manually, which is a labor-intensive process with no guarantee of accuracy. The dynamic and heterogeneous nature of wireless communication networks further exacerbates these challenges. The primary hurdle lies in the unprecedentedly enormous and ever-expanding scale of wireless networks. The resultant WDKG includes a vast array of fields and relations, the number of which are both on the order of thousands. Moreover, heterogeneity abounds in the various attributes of nodes (e.g., \emph{block error rate} from the physical layer and \emph{average throughput} from the MAC layer) as listed in Appendix, alongside multiple types of edges (e.g., causal/explicit/implicit relations). Such extensive heterogeneity complicates manual differentiation and often prone to errors. Additionally, the edges of WDKGs keeps evolving with scene variation (e.g., the mobility of transmitter and receiver), which necessitates near real-time tracking, analysis, and updating. Given these complexities, the development of an automated technique for constructing and refining WDKGs becomes imperative.

Graph structure learning (GSL) methods \cite{Chen2020, Francechi2019} enable automatic topology construction but traditionally rely on labels for supervision, resulting in biased structures due to the neglect of unlabeled nodes or edges, which limits scalability. To address this issue, self-supervision GSL paradigms have emerged, which leverages supervision signals from contrastive \cite{Liu2022} or generative learning \cite{Fatemi2021}.  Nevertheless, these approaches are primarily tailored for static homogeneous graphs such as Cora, CiteSeer and PubMed, presenting challenges for structure learning of dynamic and heterogeneous WDKGs. Motivated by these challenges, we propose a novel GSL paradigm that integrates attention mechanisms into self-supervised GSL. To capture heterogeneous information and evolutionary patterns between consecutive snapshots, we employ a hierarchical attention model and temporal attention model. Our contributions are summarized as follows.

(1) We pioneer GSL on WDKGs. Specifically, we slice the wireless networks based on edge properties and segmenting the time-varying network into static ones using coherence time. This enables automatic construction and refinement of network topology, reducing labor costs and subjective biases associated with manual construction, while enhancing network automation efficiency.

(2) We propose a novel unsupervised data-and-model driven graph structure learning (DMGSL) method. The hierarchical attention module integrates the sliced WDKG with single-edge attributes, while the temporal attention module incorporates historical structures with the current structure using weighted contribution factors. This allows for learning dynamic and heterogeneous WDKG at a finer granularity.

(3) We apply the proposed method to the real WDKG dataset and evaluate its performance on the node classification task. The results verify that our method outperforms several state-of-the-arts in terms of WDKG structure refinement and the classification performance, as measured by accuracy, precision, recall, and F1-score.\footnote{The link to the dataset and codes will be made available online upon acceptance, adhering to the double-blind review policy.}

\section{Related Work}
\subsection{Graph Structure Learning}

Recent years has witnessed a growing interest in learning graph structures for graph neural networks (GNNs) by modeling the adjacency matrix with learnable parameters and optimizing them alongside GNNs for downstream tasks. Approaches to parameterize the adjacency matrix can be broadly categorized into three types. The first type is model-based \cite{Wang2021, Francechi2019}, where the discrete nature of graph structures is taken into account by modeling them as probabilistic models such as Bernoulli and stochastic block models. The second type is based on the similarity matrix \cite{Chen2020, Yu2021}, where node similarities are evaluated using various metric functions such as cosine similarity and dot product. The third type treats each entry of the adjacency matrix as a directly learnable parameter \cite{Jin2020}. However, these GSL methods heavily hinge on labeled data for supervision, which can yield biased structures as the learning process prioritizes labeled nodes and edges.

\subsection{Self-supervision Learning}
To extend the applicability of GSL to semi-supervised and unsupervised contexts, self-supervision has emerged. Self-supervision falls into two main categories: generative methods and contrastive methods. The former concentrates on minimizing the reconstruction error, typically achieved through autoencoder \cite{Zhu2020,Huang2022}, which aim to preserve essential information of the original data at a pixel-level. Contrastive methods, taking a different approach, aim to train models capable of effectively distinguishing different inputs in the feature space. For instance, Liu et al. \shortcite{Liu2022} employ self-supervision via muti-view graph contrastive learning, where the mutual information between the anchor view and the learned view is maximized. In comparison to reconstructing the original data, the latter approach is more tractable and scalable.

\subsection{Dynamic Heterogeneous Graphs Learning}
The real-world graphs usually exhibit dynamic and heterogeneous characteristics. In general, the dynamic graphs can be modeled as snapshot sequences and timestamp graphs. Since the timestamp model can be transformed into a snapshot model with an appropriate granularity, research methods based on the snapshot model are more adaptable and will be the focus of our discussion. Recently there have been a multitude of researches on learning representations of dynamic heterogeneous graphs. There are two main types of approaches. The first type is the incremental method, which leverages the embedding of the last snapshot to learn the current embedding \cite{Wang2022}. This method is computationally efficient but suffers from error accumulation and can only capture short-term temporal information. The second type is the retrained method, which learns embeddings for each snapshot and designs neural networks to capture temporal information \cite{Yang2020}. This approach can capture long-term temporal information but becomes computationally intricate as the number of timesteps increases. To address the computational challenges, self-attention has emerged to selectively learn the most relevant historical information and disregard unnecessary information \cite{Sankar2020}. These methods have been verified efficient in the embedding learning for dynamic heterogeneous graphs, but few researches have focused on the task of structure learning in this context.

\begin{figure*}[ht]
\centering
\includegraphics[width=0.9\textwidth]{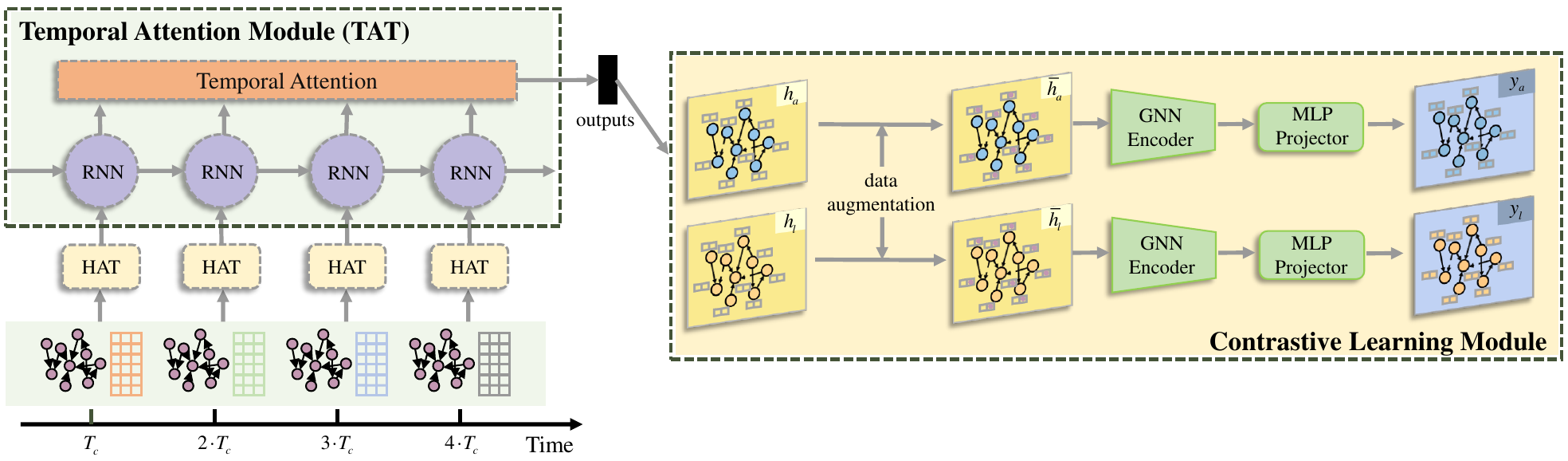} 
\caption{Overall architecture of the proposed DMGSL. It consists of three modules: a) Hierarchical attention module (HAT). The input is the adjacency matrix of expert knowledge and feature matrix, the output is the anchor graph and learned graph integrating different types of edge; b) Temporal attention module (TAT). The input is the output of a) at different snapshots, the output is the anchor graph and learned graph integrating temporal information; c) Contrastive learning module.  Calculate the contrastive loss after encoding and mapping the output of b), providing a self-supervised signal for unsupervised GSL.}
\label{fig1}
\end{figure*}

\section{Problem Definition}

Before the specific statement of our method, we first make a definition of WDKG. As mentioned before, the methods for learning dynamic heterogeneous graphs are divided into two main streams. In this paper, the snapshots method is adopted to fit the concept of coherence time in communication network. The coherence time ${{T}_{c}}$ is introduced for dividing dynamic graphs, during which the channel can be reasonably viewed as time-invariant. As a result, the dynamic heterogeneous WDKG can be viewed as a series of static heterogeneous snapshots, denoted as  $\mathbb{G}=\left\{ {{\mathcal{G}}^{t\cdot {{T}_{c}}}}|t=1,2,\cdots ,T \right\}$, where $T$ is the number of snapshots. Each snapshot ${{\mathcal{G}}^{t}}=\left({{\mathbf{X}}^{t}}, \mathcal{V},{{\mathcal{E}}^{t}} \right)=\left( {{\mathbf{X}}^{t}},{{\mathbf{A}}^{t}} \right)$ ($T_{c}$ is omitted for convenience, so as in the following of this paper) is a static heterogeneous graph where $\mathcal{V}$ is a shared node set and $n=\left| \mathcal{V} \right|$ represents the number of nodes, ${\mathcal{E}^{t}}$ represents edge set and $m=\left| {{\mathcal{E}}^{t}} \right|$ is the number of edges at time step $t$, ${{\mathbf{X}}^{t}}\in {{\mathbb{R}}^{n\times d}}$ is the node feature matrix at time $t$ (the $i$-th row $x_{i}^{t}$ is the feature vector of node ${{v}_{i}}$ at time step $t$), ${{\mathbf{A}}^{t}}\in {{\left[ 0,1 \right]}^{n\times n}}$ is the adjacency matrix (where ${{a}_{ij}}$ is the weight of the edge from ${{v}_{i}}$ to ${{v}_{j}}$ at time step $t$). Considering the heterogeneity of each snapshot, ${{\mathbf{A}}^{t}}=\left\{ \mathbf{A}_{1}^{t},...,\mathbf{A}_{s}^{t} \right\}$, where $s$ is the number of edge categories. Given a dynamic heterogeneous graph $\mathbb{G}$, our target is to refine the graph structure based on the existing graph structure and feature matrix.

\section{Methodology}
In this section, the proposed framework of DMGSL will be explained in detail, the framework is depicted in Fig. \ref{fig1}.

\subsection{Hierarchical attention module}

\begin{figure*}[ht]
\centering
\includegraphics[width=0.88\textwidth]{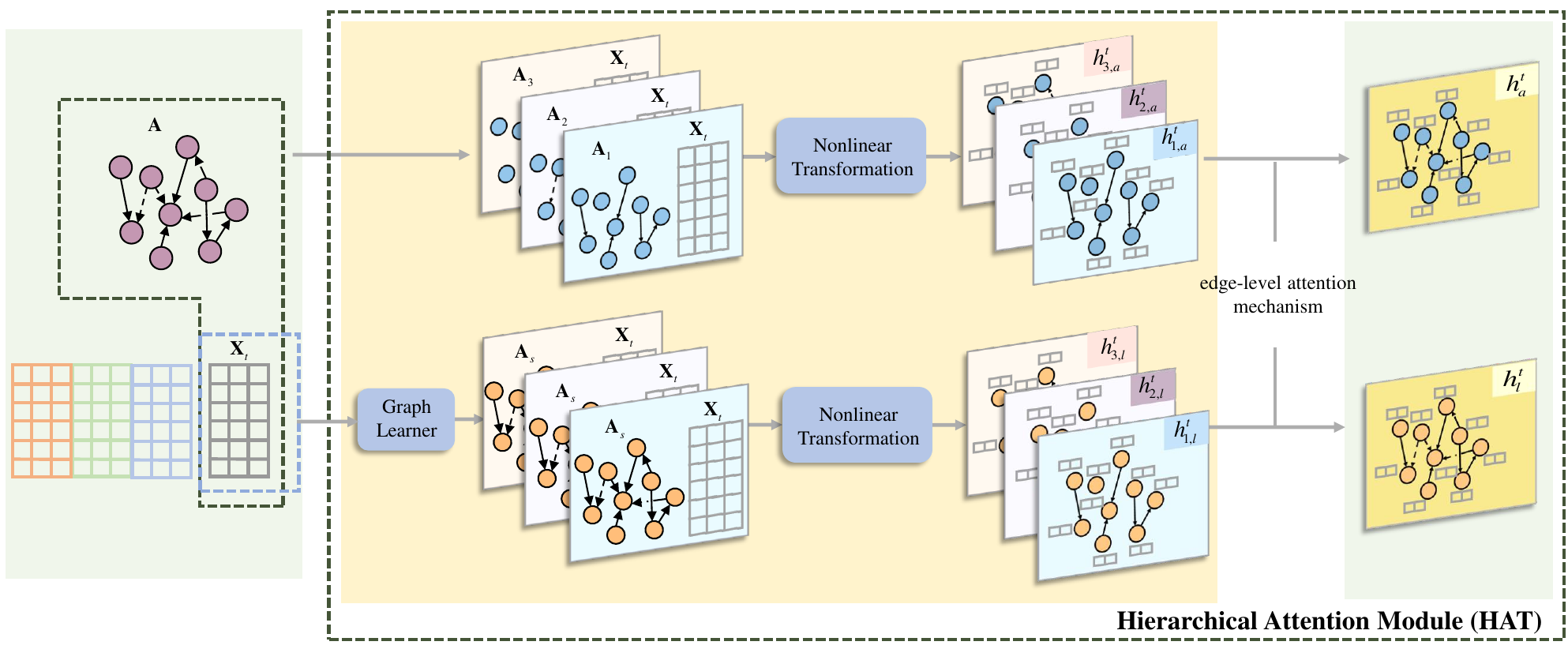} 
\caption{Schematic diagram of hierarchical attention module. The input is the adjacency matrix constructed by experts and the feature matrix of one snapshot. In the upper branch, the adjacency matrix is divided into three sub-adjacency matrices (causal/implicit/explicit relations) and mapped to a new feature space along with feature matrix. In the lower branch, the adjacency matrix is learned by a graph learner and then mapped to a new feature space along with feature matrix. The three anchor graphs and learned graphs are fused respectively through the edge-level attention mechanism.}
\label{fig2}
\end{figure*}

Given the heterogeneous nature of the wireless network, it is sliced in accordance with various relations (or edges). In terms of the dataset we use in this paper, three distinct sliced sub-networks are acquired. Then a hierarchical attention module is devised to independently learn each slice and subsequently merge them using an attention mechanism, which facilitates a nuanced understanding of the discrepancy in relations, allowing for fine-grained learning. The diagram of hierarchical attention module (HAT) is shown in Fig. \ref{fig2}. 

\subsubsection{Anchor graphs and learned graphs}

As defined before, the dynamic heterogeneous WDKG is denoted as a series of static snapshots $\mathbb{G}=\left\{ {{\mathcal{G}}^{1}},{{\mathcal{G}}^{2}},...,{{\mathcal{G}}^{T}} \right\}$. As to the static graph at $t$-th snapshot ${{\mathcal{G}}^{t}}$, the adjacency matrix ${{\mathbf{A}}^{t}}$ is divided into three sub-adjacency matrices according to different kinds of  relations (or edges), i.e., $\mathbf{A}_{1}^{t}$, $\mathbf{A}_{2}^{t}$, $\mathbf{A}_{3}^{t}$. Combined with ${{\mathbf{X}}^{t}}$ (the feature matrix at time $t$), we get $\mathbf{E}_{1,a}^{t}=\left( {{\mathbf{X}}^{t}},\mathbf{A}_{1}^{t} \right)$, $\mathbf{E}_{2,a}^{t}=\left( {{\mathbf{X}}^{t}},\mathbf{A}_{2}^{t} \right)$, $\mathbf{E}_{3,a}^{t}=\left( {{\mathbf{X}}^{t}},\mathbf{A}_{3}^{t} \right)$ as the initial matrices of anchor graphs. 

To acquire the initial matrices of learned graphs, a full graph parameterization (FPG) learner is considered to generate sketchy adjacency matrix of WDKG from feature matrix at time $t$. The FGP learner parameterizes each element of the adjacency matrix independently, the learned adjacency matrix can be presented as $\mathbf{A}_{s}^{t}$. Then $\mathbf{E}_{1,l}^{t}=\left( {{\mathbf{X}}^{t}},\mathbf{A}_{s}^{t} \right)$, $\mathbf{E}_{2,l}^{t}=\left( {{\mathbf{X}}^{t}},\mathbf{A}_{s}^{t} \right)$, $\mathbf{E}_{3,l}^{t}=\left( {{\mathbf{X}}^{t}},\mathbf{A}_{s}^{t} \right)$ are obtained as the initial matrices of learned graphs.

\subsubsection{Hierarchical attention model}

The categories of relations between fields are different, including causal, explicit and implicit relations. Therefore, we slice the wireless network into three sub-networks and learn the information of them separately. The common method is to average the learned information from three slices, but in fact these slices are of different importance to structure learning. For instance, the direct influence of the causal relation (or edge) on the structure is higher than the indirect influence of the implicit relation (or edge). Therefore, we introduce a hierarchical attention model to learn the importance of different edges to GSL of wireless network, so as to integrate the information of the three network slices more explainably.

Specifically, the initial matrices of anchor graphs and learned graphs (i.e. $\mathbf{E}_{1,a}^{t}$, $\mathbf{E}_{2,a}^{t}$, $\mathbf{E}_{2,a}^{t}$, $\mathbf{E}_{1,l}^{t}$, $\mathbf{E}_{2,l}^{t}$, $\mathbf{E}_{3,l}^{t}$) are firstly entered into a nonlinear transformation function so that they map to the same feature space by $\sigma \left( \mathbf{W}\cdot \mathbf{e}_{s,a}^{t}+\mathbf{b} \right)$, $\sigma \left( \mathbf{W}\cdot \mathbf{e}_{s,l}^{t}+\mathbf{b} \right)$, where $\mathbf{e}_{s,a}^{t},e_{s,l}^{t}\in {{\mathbb{R}}^{d}}$ are the transpose of $i$-th row of initial matrices $\mathbf{E}_{s,a}^{t},\mathbf{E}_{s,l}^{t}\in {{\mathbb{R}}^{n\times (d+n)}}$ ($i$ is omitted for convenience) which are two matrix representations of WDKG, $\sigma $ denotes the activation function, $\mathbf{W}$ and $\mathbf{b}$ represent the weight matrix and bias vector, parameters of which are shared by anchor and learned graphs in the same edge-level (e.g., $\mathbf{E}_{1,a}^{t}$ and $\mathbf{E}_{1,l}^{t}$ share the parameters of $\mathbf{W}$ and $\mathbf{b}$). The mapped graphs are denoted as $h_{1,a}^{t}$, $h_{2,a}^{t}$, $h_{3,a}^{t}$ and $h_{1,l}^{t}$, $h_{2,l}^{t}$, $h_{3,l}^{t}$. Then, the similarities between the mapped graphs and the edge-level attention parameterized vector are calculated to evaluate the importance of different slices to structure learning. Sequently, the normalized weight factors of the anchor graphs and learned graphs with edge type $s$ at time $t$ are figured up, which are denoted as $\alpha _{s,a}^{t}$, $\alpha _{s,l}^{t}$. The process can be defined as:
\begin{equation}
\begin{aligned}
	& \alpha _{s,a}^{t}=\frac{\exp \left( {{\mathbf{q}}^{\mathsf{T}}}\cdot \sigma \left( \mathbf{W}\cdot \mathbf{e}_{s,a}^{t}+\mathbf{b} \right) \right)}{\sum\nolimits_{s}{\exp \left( {{\mathbf{q}}^{\mathsf{T}}}\cdot \sigma \left( \mathbf{W}\cdot \mathbf{e}_{s,a}^{t}+\mathbf{b} \right) \right)}}, s\in [1,m] \\ 
	& \alpha _{s,l}^{t}=\frac{\exp \left( {{\mathbf{q}}^{\mathsf{T}}}\cdot \sigma \left( \mathbf{W}\cdot \mathbf{e}_{s,l}^{t}+\mathbf{b} \right) \right)}{\sum\nolimits_{s}{\exp \left( {{\mathbf{q}}^{\mathsf{T}}}\cdot \sigma \left( \mathbf{W}\cdot \mathbf{e}_{s,l}^{t}+\mathbf{b} \right) \right)}}, s\in [1,m]
\end{aligned}
\end{equation}

Lastly, the graphs with single communication relation can be merged with the normalized weight factors. The transpose of $i$-th row in the merged matrices $\mathbf{E}_{a}^{t},\mathbf{E}_{l}^{t}\in {{\mathbb{R}}^{n\times (d+n)}}$ are expressed as $\mathbf{e}_{a}^{t}$ and $\mathbf{e}_{l}^{t}$, which can be viewed as two kinds of embeddings of node $i$ in the wireless communication network at time $t$. The merge can be formulized as:
\begin{equation}
\begin{aligned}
	& \mathbf{e}_{a}^{t}=\sum\limits_{s=1}^{m}{\alpha _{s,a}^{t}\cdot \mathbf{e}_{s,a}^{t}},
	& \mathbf{e}_{l}^{t}=\sum\limits_{s=1}^{m}{\alpha _{s,l}^{t}\cdot \mathbf{e}_{s,l}^{t}}. \\ 
\end{aligned}
\end{equation} 
By performing the above operations on the WDKG at each time step, the anchor graphs and the learned graphs of each snapshot can be obtained, denoted as $h_{a}^{t}$ and $h_{l}^{t}$, the corresponding matrices are represented as $\left\{ \mathbf{E}_{a}^{1},\mathbf{E}_{a}^{2},\cdots ,\mathbf{E}_{a}^{T}\in {{\mathbb{R}}^{n\times D}} \right\}$, $\left\{ \mathbf{E}_{l}^{1},\mathbf{E}_{l}^{2},\cdots ,\mathbf{E}_{l}^{T}\in {{\mathbb{R}}^{n\times D}} \right\}$, where $n$ is the number of nodes and $D$ is the output dimension of  hierarchical attention model.

\subsection{Temporal Attention Module}

The wireless communication system encompasses a highly complex channel, including a series of channels caused by obstacles such as reflection, diffraction, scattering, coherence and shadowing. The mobility of transmitter or receiver renders time-invariance elusive. To address this, the notion of coherence time is introduced, representing the duration in which the channel can reasonably be viewed as time-invariant. Assuming the transmitter is fixed, the coherence time can be calculated based on the radio frequency and the radial velocity of receiver. Consequently, the dynamic WDKG is partitioned into several static snapshots accordingly. On this basis, the temporal attention module (TAT) is introduced to effectively integrate the abundant temporal features inherent in the WDKG.

\subsubsection{Long short-term memory unit}

In this paper, to model the dynamic information of WDKG, we employ a basic variant of RNN called long short-term memory (LSTM)  which is qualified for conveying information during a long time. To cater to the limited memory units, LSTM preserves useful information that needs long-term memory, and forgets superfluous information. Moreover, a mechanism that can dynamically adjust memory is introduced to update the valuable information that needs to be remembered in time. Specifically, the LSTM model contains state vector ${{\mathbf{s}}^{t}}$, forgetting vector ${{\mathbf{f}}^{t}}$, memory vector ${{\mathbf{c}}^{t}}$, input vector ${{\mathbf{i}}^{t}}$, output vector ${{\mathbf{o}}^{t}}$. The matrices output from hierarchical attention model are represented as $\left\{ \mathbf{E}_{a}^{1},\mathbf{E}_{a}^{2},\cdots ,\mathbf{E}_{a}^{T}\in {{\mathbb{R}}^{n\times D}} \right\}$, $\left\{ \mathbf{E}_{l}^{1},\mathbf{E}_{l}^{2},\cdots , \mathbf{E}_{l}^{T}\in {{\mathbb{R}}^{n\times D}} \right\}$, the transpose of their $i$-th row (denoted as $\mathbf{e}_{a}^{t},\mathbf{e}_{l}^{t}\in {{\mathbb{R}}^{D}}$, where $i$ is omitted) are entered into LSTM. An LSTM unit can be represented by the following formulas (omit the subscripts of anchor and learned graphs):
\begin{equation}
\begin{aligned}
	& {{\mathbf{i}}^{t}}=\sigma \left( {{\mathbf{W}}_{i}}\cdot \left[ {{\mathbf{e}}^{t}}\parallel {{\mathbf{s}}^{t-1}} \right]+{{\mathbf{b}}_{i}} \right), \\
	& {{\mathbf{f}}^{t}}=\sigma \left( {{\mathbf{W}}_{f}}\cdot \left[ {{\mathbf{e}}^{t}}\parallel {{\mathbf{s}}^{t-1}} \right]+{{\mathbf{b}}_{f}} \right), \\ 
	& {{\mathbf{o}}^{t}}=\sigma \left( {{\mathbf{W}}_{o}}\cdot \left[ {{\mathbf{e}}^{t}}\parallel {{\mathbf{s}}^{t-1}} \right]+{{\mathbf{b}}_{o}} \right), \\ 
	& {{{\mathbf{\tilde{c}}}}^{t}}=\tanh \left( {{\mathbf{W}}_{c}}\cdot \left[ {{\mathbf{e}}^{t}}\parallel {{\mathbf{s}}^{t-1}} \right]+{{\mathbf{b}}_{c}} \right), \\ 
	& {{\mathbf{c}}^{t}}={{\mathbf{f}}^{t}}\odot {{\mathbf{c}}^{t-1}}+{{\mathbf{i}}^{t}}\odot {{{\mathbf{\tilde{c}}}}^{t}}, 
	 {{\mathbf{s}}^{t}}={{\mathbf{o}}^{t}}\odot \tanh \left( {{\mathbf{c}}^{t}} \right), \\ 
\end{aligned}
\end{equation}
where $t\in \left\{ 1, 2,\cdots, T \right\}$, ${{\mathbf{i}}^{t}}$, ${{\mathbf{f}}^{t}}$, ${{\mathbf{o}}^{t}}$, ${{\mathbf{c}}^{t}}$ $\in$ ${{\mathbb{R}}^{F}}$ ($F$ is the output dimension of LSTM), ${{\mathbf{W}}_{i}}$, ${{\mathbf{W}}_{f}}$, ${{\mathbf{W}}_{o}}$, ${{\mathbf{W}}_{c}}$ $\in$ ${{\mathbb{R}}^{F\times 2D}}$ and ${{\mathbf{b}}_{i}}$, ${{\mathbf{b}}_{f}}$, ${{\mathbf{b}}_{o}}$, ${{\mathbf{b}}_{c}}$ $\in$ ${{\mathbb{R}}^{F}}$ are trainable parameters, $\sigma$ is the activation function, $\parallel $ is the concatenation operation, $\odot $ is the Hadamard product. ${{\mathbf{s}}^{0}}$, ${{\mathbf{c}}^{0}}$ $\in$ ${{\mathbb{R}}^{D}}$ need to be initialized, and in this article they are initialized to an all-one vector $\boldsymbol{1}$. The outputs of LSTM (i.e., the graph structure matrix at $t$ time step that has integrated historical information) can be defined as ${{\mathbf{S}}^{t}}={{\left[ {{\left( \mathbf{s}_{1}^{t} \right)}^{\mathsf{T}}},{{\left( \mathbf{s}_{2}^{t} \right)}^{\mathsf{T}}},\cdots ,{{\left( \mathbf{s}_{n}^{t} \right)}^{\mathsf{T}}} \right]}^{\mathsf{T}}}$ $\in$ ${{\mathbb{R}}^{n\times F}}$.

\subsubsection{Temporal attention model}

To fuse the acquired wireless network topology across different snapshots, we employ a temporal attention model. This model calculates contribution factors to determine the impact of various snapshots on the overall structure learning process. For example, snapshots with lower doppler effect (i.e., a lower radial velocity between transmitter and receiver) may be more significant in the learning of network topology compared to other snapshots. These contribution factors enable us to effectively weigh the influence of each snapshot on the overall structure learning process. The $i$-th row of the graph structure at all times are extracted to constitute a fresh matrix ${{\mathbf{S}}_{i}}={{\left[ {{\left( \mathbf{s}_{i}^{1} \right)}^{\mathsf{T}}},{{\left( \mathbf{s}_{i}^{2} \right)}^{\mathsf{T}}},\cdots ,{{\left( \mathbf{s}_{i}^{T} \right)}^{\mathsf{T}}} \right]}^{\mathsf{T}}},\mathbf{s}_{i}^{1},\mathbf{s}_{i}^{2},\cdots ,\mathbf{s}_{i}^{T}\in {{\mathbb{R}}^{F}}$, which is the input. The popular scaling dot multiplication attention mechanism in natural language processing is adopted. The input ${{\mathbf{S}}_{i}}\in {{\mathbb{R}}^{T\times F}}$ is multiplied with three parameter matrices ${{\mathbf{W}}^{Q}}$, ${{\mathbf{W}}^{K}}$, ${{\mathbf{W}}^{V}}$ $\in$ ${{\mathbb{R}}^{F\times {F}^{'}}}$, mapping into different feature spaces, represented as $\mathbf{Q}$, $\mathbf{K}$, $\mathbf{V}$ $\in$ ${{\mathbb{R}}^{T\times {F}^{'}}}$, which is viewed as the linear transformation of the input. Using $\mathbf{Q}$, $\mathbf{K}$, $\mathbf{V}$ rather than ${{\mathbf{S}}_{i}}\in {{\mathbb{R}}^{T\times F}}$ in the calculation enhances the fitting ability of the model effectively. Then, multiply $\mathbf{Q}$ and ${{\mathbf{K}}^{\mathsf{T}}}$ to generate the similarity matrix. What needs to be emphasized is that when the dimension of $\mathbf{K}$ increases, the variance of $\mathbf{Q}\cdot {{\mathbf{K}}^{\mathsf{T}}}$ will become larger. In order to reduce the variance, each element of the similarity matrix is divided by $\sqrt{{{F}^{'}}}$ (${{F}^{'}}$ is the dimension of $\mathbf{K}$. The normalized similarity matrix can be regarded as a weight matrix. Finally, multiply the weight matrix with $\mathbf{V}$ and calculate the weighted sum, which is the output of the network. The above process can be formulized as follows:
\begin{equation}
\begin{aligned}
	{{\mathbf{Z}}_{i}} & ={{\Gamma }_{i}}\cdot {{\mathbf{V}}_{i}}=\text{softmax}\left( \frac{ {{\mathbf{Q}}}\cdot{\mathbf{K}^{\mathsf{T}}}}{\sqrt{{{F}^{'}}}}+\mathbf{M} \right)\cdot{\mathbf{V}},
\end{aligned}
\end{equation}
where  ${{\Gamma }_{i}}\in {{\mathbb{R}}^{T\times T}}$ is the weight matrix, $\mathbf{M}\in {{\mathbb{R}}^{T\times T}}$ is the masking matrix. If ${{M}_{u\eta}}=-\infty $, there is no effect from time $u$ to $\eta$, and the corresponding element in the weight matrix is 0. If $u$ is earlier than $\eta$, then ${{M}_{u\eta}}=0$; Otherwise, ${{M}_{u\eta}}=-\infty $. The output of the temporal attention model is defined as ${{\mathbf{Z}}_{i}}={{\left[ {{\left( \mathbf{z}_{i}^{1} \right)}^{\mathsf{T}}},{{\left( \mathbf{z}_{i}^{2} \right)}^{\mathsf{T}}},\cdots ,{{\left( \mathbf{z}_{i}^{T} \right)}^{\mathsf{T}}} \right]}^{\mathsf{T}}},\mathbf{z}_{i}^{1},\mathbf{z}_{i}^{2},\cdots ,\mathbf{z}_{i}^{T}\in {{\mathbb{R}}^{{{F}^{'}}}}$, where ${{F}^{'}}$ is the output dimension. 

In order to enhance the performance, the extended multi-head attention mechanism is used. To be concrete, we define multiple groups (i.e. $\kappa $ groups) of ${{\mathbf{W}}^{Q}}$, ${{\mathbf{W}}^{K}}$, ${{\mathbf{W}}^{V}}$, each group is calculated separately to generate different $\mathbf{Q}$, $\mathbf{K}$, $\mathbf{V}$,  and learn various parameters, the obtained multiple outputs are concatenated to get ${{\mathbf{Z}}_{i}}=\text{Concat}\left( \mathbf{\hat{Z}}_{i}^{1},\mathbf{\hat{Z}}_{i}^{2},\cdots ,\mathbf{\hat{Z}}_{i}^{\kappa } \right)$. In this paper, we take all the $\mathbf{z}_{i}^{T}$ in ${{\mathbf{Z}}_{i}}$ to constituent matrix $\mathbf{E}={{\left[ {{\left( \mathbf{z}_{1}^{T} \right)}^{\mathsf{T}}},{{\left( \mathbf{z}_{2}^{T} \right)}^{\mathsf{T}}},\cdots ,{{\left( \mathbf{z}_{n}^{T} \right)}^{\mathsf{T}}} \right]}^{\mathsf{T}}}\in {{\mathbb{R}}^{n\times {{F}^{'}}}}$ as output. The anchor graph and learned graph formed from temporal attention module can be denoted as $h_{a}$ and $h_{l}$ which are regard as two graph representations of the wireless network.

\subsection{Contrastive learning module}

Data augmentation is significant method to mitigate overfitting and explore richer information. We use two common data augmentation schemes, edge dropping and feature masking, the augmented matrices $\mathbf{\bar{E}}_{a}$ and $\mathbf{\bar{E}}_{l}$ are obtained, corresponding to the two augmented views $\bar{h}_{a}$ and $\bar{h}_{l}$.

Next, the augmented graphs are encoded and compressed, transforming the high-dimension into lower dimension. Graph convolutional network (GCN) is exploited as the encoder, the encoding process can be expressed as
\begin{equation}
\begin{aligned}
	& \mathbf{H}_{a}=\text{GC}{{\text{N}}_{\theta }}\left( \mathbf{\bar{E}}_{a} \right), \mathbf{H}_{l}=\text{GC}{{\text{N}}_{\theta }}\left( \mathbf{\bar{E}}_{l} \right), s\in [1,m],
\end{aligned}
\end{equation}
where $\theta $ is the parameter of GCN encoder, $\mathbf{H}_{a}$, $\mathbf{H}_{l}$ $\in$ ${{\mathbb{R}}^{n\times {{d}_{1}}}}$ (${{d}_{1}}$ represents the output dimension of encoder) are the encoded structure matrices.

\begin{figure*}[!h]
	\centering
	\subfigure[Raw]{
		\includegraphics[scale=0.33]{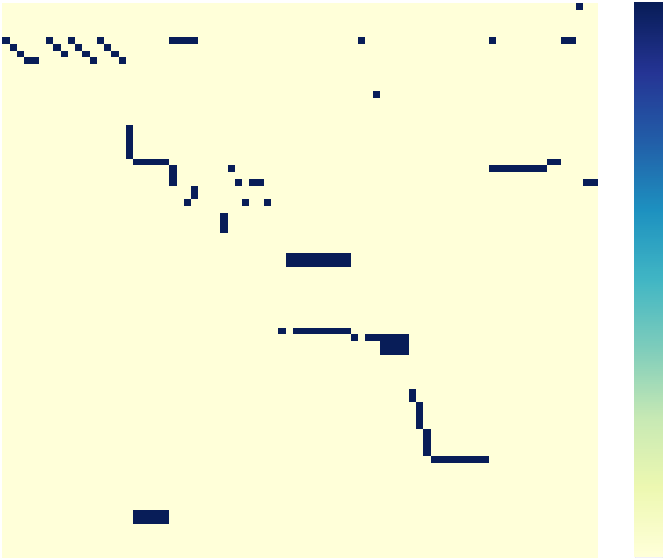}}\hfil
	\subfigure[IDGL]{
		\includegraphics[scale=0.33]{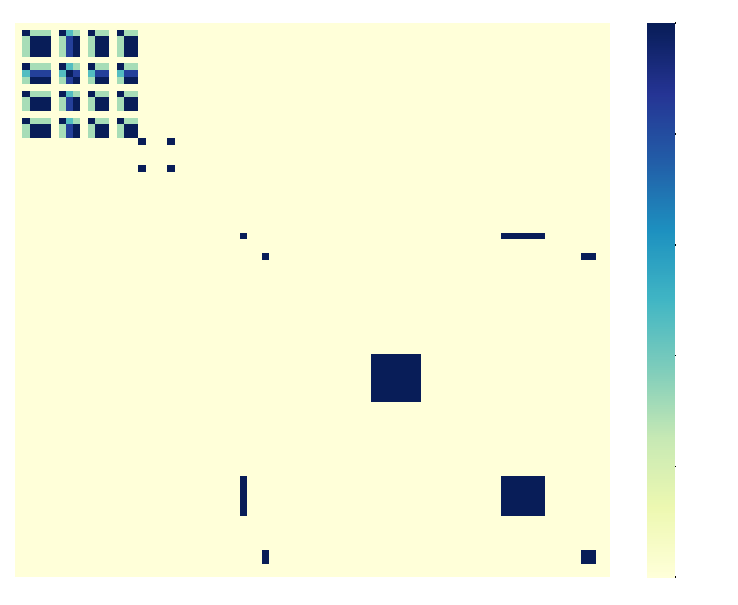}}\hfil
	\subfigure[SLAPS]{
		\includegraphics[scale=0.33]{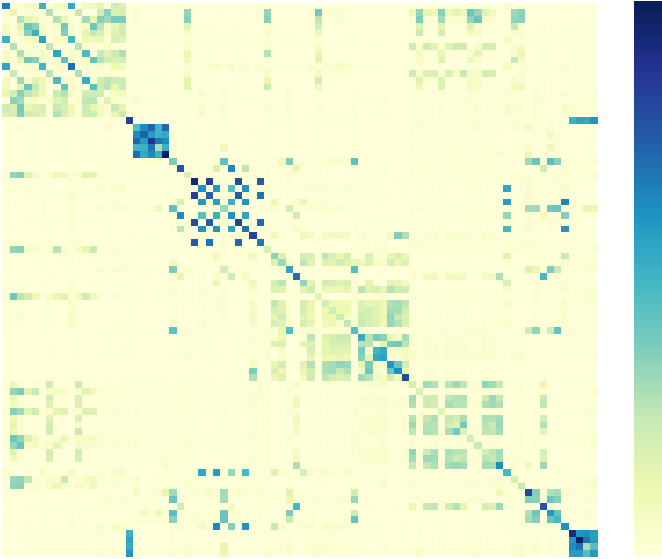}}\hfil
	\subfigure[Sublime]{
		\includegraphics[scale=0.33]{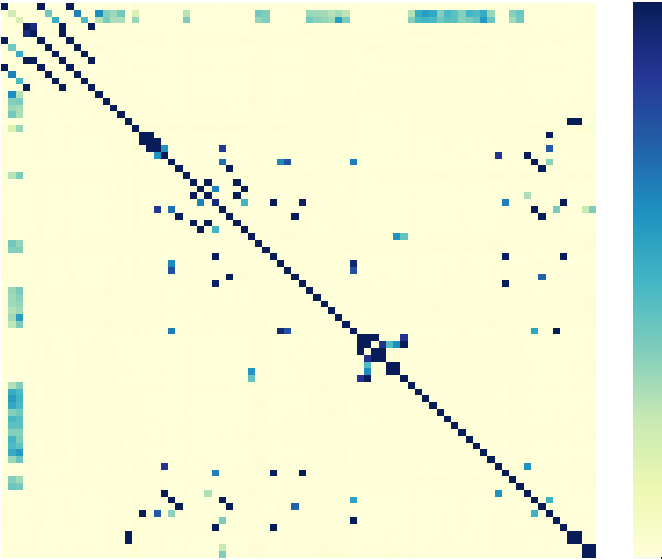}}\hfil
	\subfigure[GEN]{
		\includegraphics[scale=0.33]{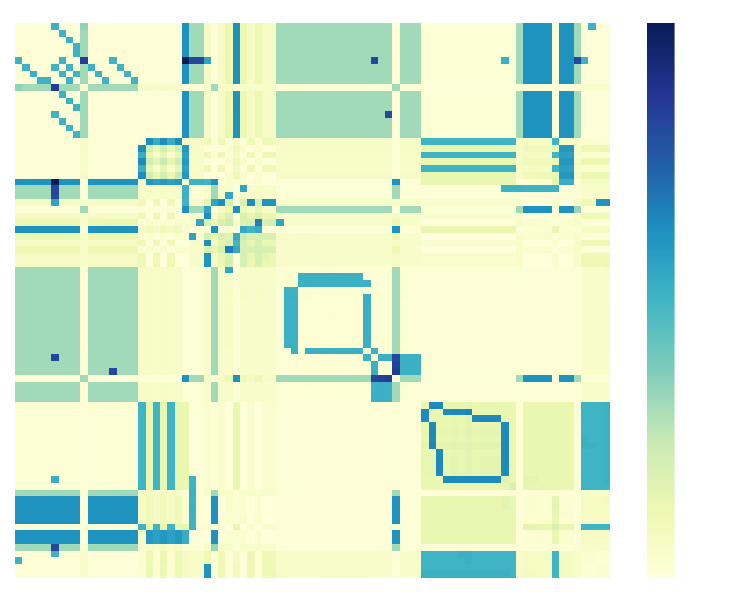}}\hfil
	\subfigure[IDGL-Anch]{
		\includegraphics[scale=0.33]{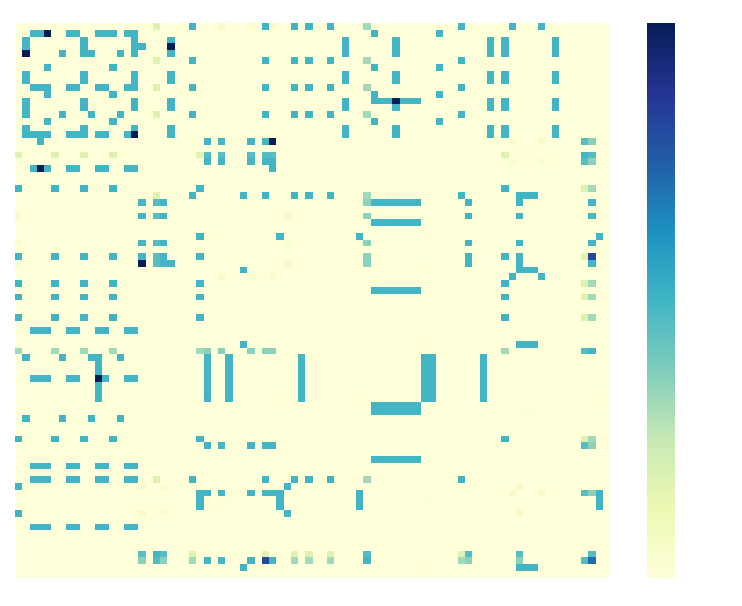}}\hfil
	\subfigure[SLAPS-2s]{
		\includegraphics[scale=0.33]{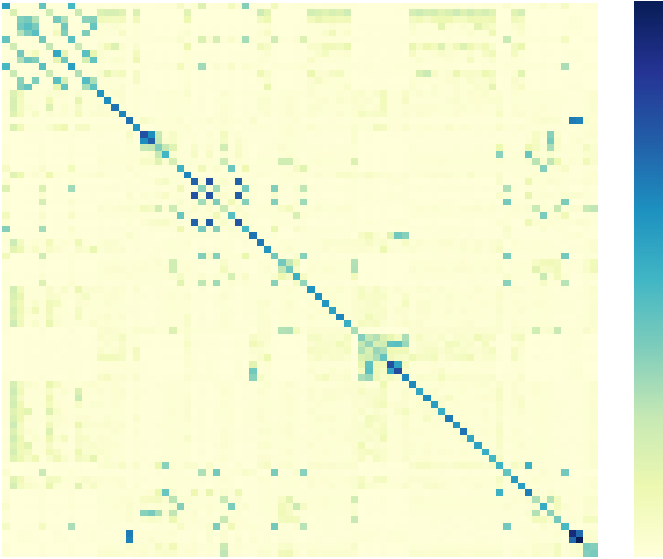}}\hfil
	\subfigure[DMGSL]{
		\includegraphics[scale=0.33]{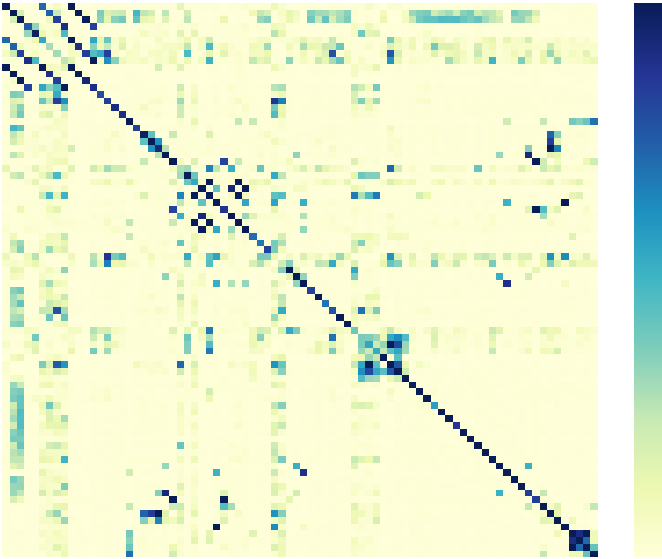}}\hfil
	\caption{Heatmaps of adjacency matrices.}
	\label{fig3}
\end{figure*}

Furthermore, to calculate the contrast loss function,  we reflect the views to another latent space with the assistance of multiple layer projection (MLP), which is formalize as:
\begin{equation}
\begin{aligned}
	{{{\mathbf{Y}}_{a}}={{g}_{\varphi }}\left( {{\mathbf{H}}_{a}} \right),{{\mathbf{Y}}_{l}}={{g}_{\varphi }}\left( {{\mathbf{H}}_{l}} \right)}
\end{aligned}
\end{equation}
where $\varphi $ is the parameter of the projector ${{g}_{\varphi }}\left( \cdot  \right)$, ${{\mathbf{Y}}_{a}},{{\mathbf{Y}}_{l}}\in {{\mathbb{R}}^{n\times {{d}_{2}}}}$ (${{d}_{2}}$ is the output dimension of projector) are mapped graph matrices of the anchor and learned graphs. Then, a contrast learning loss function \cite{Aaron2019} is used to maximize the similarity between each row vector of the two graph structures:
\begin{align}
& {{\mathcal{L}}}=\frac{1}{2n}\sum\limits_{i=1}^{n}{\left[ \ell \left( {{\mathbf{y}}_{a,i}},{{\mathbf{y}}_{l,i}} \right)+\ell \left( {{\mathbf{y}}_{l,i}},{{\mathbf{y}}_{a,i}} \right) \right]}, \\ 
& \ell \left( {{\mathbf{y}}_{a,i}},{{\mathbf{y}}_{l,i}} \right)=\log \frac{{{e}^{\text{sim}\left( {{\mathbf{y}}_{a,i}},{{\mathbf{y}}_{l,i}} \right)/p}}}{\sum\nolimits_{k=1}^{n}{{{e}^{\text{sim}\left( {{\mathbf{y}}_{a,i}},{{\mathbf{y}}_{l,i}} \right)/p}}}},
\end{align}
where  $\text{sim}\left( \cdot ,\cdot  \right)$ is the cosine similarity function, $p$ is the temperature parameter. The loss function is minimized during the training process, where involves iteratively updating the model's parameters using gradient descent. 

\begin{table*}[!h]
	\renewcommand\arraystretch{1.2}
	\begin{center}
		\begin{small}
			\begin{tabular}{cccccc}
				\toprule
				\textbf{Dataset} & \textbf{Method} & \textbf{Accuracy} & \textbf{Precision} & \textbf{Recall} & \textbf{F1-score} \\
				\midrule
				& IDGL & 0.4375 +/- 0.1046 & 0.3515 +/- 0.1467 & 0.4375 +/- 0.1046 & 0.373 6+/-0.1343 \\
				& IDGL-Anch & 0.3625 +/-0.1000 & 0.2221 +/-0.1367 & 0.3625 +/- 0.1000 & 0.2571 +/- 0.1332 \\
				& SLAPS & 0.5875 +/- 0.0637 & 0.4764 +/- 0.1268 & 0.5875 +/- 0.0637 & 0.5100 +/- 0.0974 \\
				Uplink throughput & SLAPS-2s & 0.6000 +/- 0.0637 & 0.5525 +/- 0.0788 & 0.6000 +/- 0.0729 & 0.5224 +/- 0.0766 \\
				data (35min) & GEN & 0.5750 +/- 0.0729 & 0.5439 +/- 0.0723 & 0.5750 +/- 0.0729 & 0.5302 +/- 0.0632 \\
				& Sublime & 0.6125 +/- 0.0468 & 0.5112 +/- 0.0644 & 0.6125 +/- 0.0468 & 0.5438 +/- 0.0550 \\
				&DMGSL (w/o TAT) & 0.6125 +/- 0.0250 & 0.5639 +/- 0.0204 & 0.6125 +/- 0.0250 & 0.5538 +/- 0.0356 \\
				&DMGSL (w/o HAT) & \underline{0.6625 +/- 0.0306} & \underline{0.6108 +/- 0.0417} & \underline{0.6625 +/- 0.0306} & \underline{0.6104 +/- 0.0350} \\
				&DMGSL & \textbf{0.7000 +/- 0.0250} & \textbf{0.6373 +/- 0.0321} & \textbf{0.7000 +/- 0.0250} & \textbf{0.6436 +/- 0.0321} \\
				\bottomrule
			\end{tabular}
		\end{small}
	\end{center}
	\caption{Performance of node classification (values with standard deviation). The highest is highlighted with \textbf{boldface}, the second highest is highlighted with \underline{underline}.}
	\label{Table3}
\end{table*}

\section{Experiments}

In this section, we conduct a series of experiments to examine the effectiveness of the proposed framework for fine-grained graph representation learning from the following aspects. First, we compare the learned adjacency matrix with expert knowledge and other baselines. Second, we exploit the learned structure to perform node classification tasks and compare the results with baselines. Finally, we investigate the influence of key hyperparameters on the performance of our proposed method, aiming to provide guidelines for parameter tuning. The detailed setup of experiments can be found in Appendix~A.

\subsection{Baseline}

We compare the proposed method with state-of-the-art methods of structure learning, including Sublime \cite{Liu2022}, GEN \cite{Wang2021}, IDGL \cite{Chen2020}, SLAPS \cite{Fatemi2021}, and their variant, IDGL-Anch and SLAPS-2s.

\subsection{Dataset}

The dataset is collected by an automated guided vehicle which travels in the park at a speed of $10\;{\rm{km/h}}$ for $500\;{\rm{m}}$ along a planned route, receiving signals with a frequency of $3300-3800\;{\rm{Hz}}$, recording $40$ data per second. We processed and organized the data, the details of dataset we use are provided in Appendix~B.

\subsection{Comparisons}

In the proposed framework, each element of the output adjacency matrix represents the probability of an edge existing between two nodes; the higher the value, the closer the correlation between two nodes. We depict the heatmap of the original adjacency matrix and the adjacency matrices learned from the proposed method and other baselines, as shown in Fig.~\ref{fig3}. It should be noticed that the darker the patch, the more likely an edge exists between the two nodes. It can be seen that there are only a few obvious deepened color blocks in the raw adjacency matrix, which reflects limited correlations. In contrast, some baselines have learned relations concentrated near the diagonal (IDGL, SLAPS), while others are extremely scattered (IDGL-Anch). Compared to GEN, SLAPS-2s, and Sublime, the adjacency matrix learned by our proposed method presents more relations and avoids learning unnecessary relationships. To a certain extent, it modifies and refines the raw matrix effectively.

To objectively and quantitatively evaluate the effectiveness of the proposed method, we test a node classification task, whose performance is quantified with accuracy, precision, recall and F1-score. We compare the performance of the proposed DMGSL with several prevalent baselines. From Table~\ref{Table3}, we observe that DMGSL outperforms the other baselines and has clear advantages. This can be attributed to its focus on the nature of mobile communication networks. By partitioning the network hierarchically and chronologically and narrowing the learning unit, the utilization of the measured network data is maximized with hierarchical attention and temporal attention model. Other approaches, on the other hand, process all the data together, neglecting the different data fields and time-variance of mobile network, resulting in learned structures that fail to mine detailed endogenous relations.

\begin{figure}[htbp]
	\centering
	\subfigure[Accuracy versus mask rate.]{
		\includegraphics[width=0.22\textwidth]{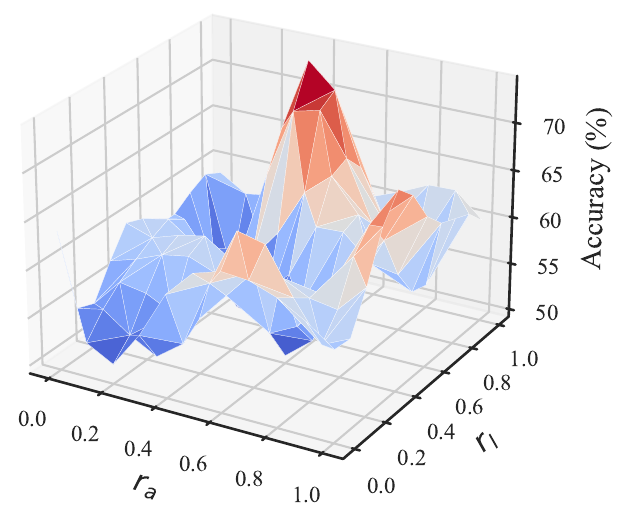}
		\label{fig6a}}
	\subfigure[Precision versus mask rate.]{
		\includegraphics[width=0.22\textwidth]{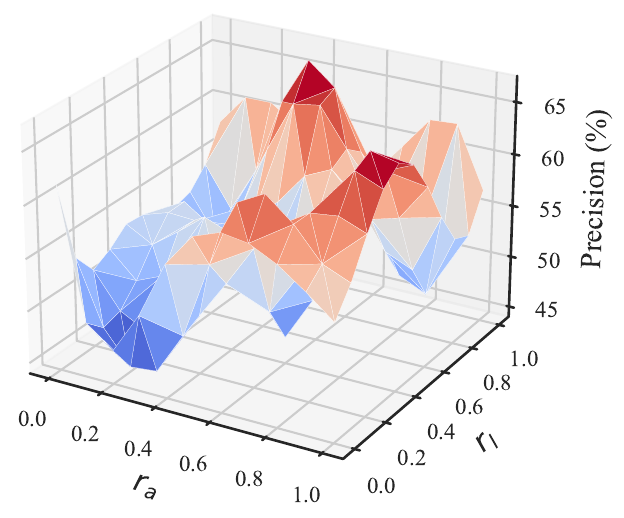}
		\label{fig6b}}
	\subfigure[Recall versus mask rate.]{
		\includegraphics[width=0.22\textwidth]{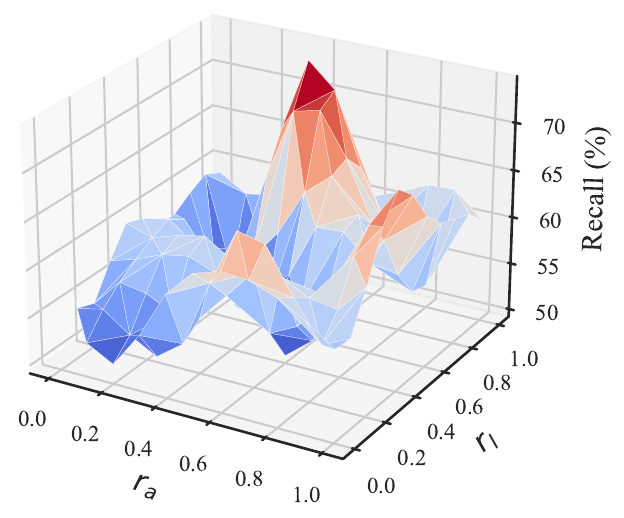}
		\label{fig6c}}
	\subfigure[F1 score versus mask rate.]{
		\includegraphics[width=0.22\textwidth]{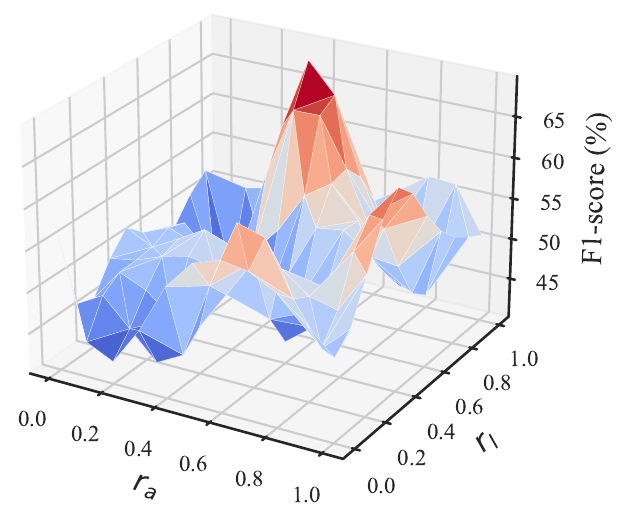}
		\label{fig6d}}
	\centering
	\caption{Impact of mask rate on the classification performance.}
	\label{fig6}
\end{figure}

\subsection{Impact of Key Hyperparameter}

Since the data fields (or nodes) and relations (or edges) of wireless communication data is less than that of general public datasets such as Cora, Citeseer, it is significant to prevent overfitting in the learning process. To this end, we perform data augmentation by feature masking before encoding and mapping. The feature masking rate of the anchor graph $r_{a}$ and learned graph $r_{l}$ are two important hyperparameters, we change these two parameters respectively in several experiments. Fig. \ref{fig6} shows the classification performance under different feature masking rate. It can be seen that both small and large $r_{a}$ correspond to poor classification performance, which is because of overfitting during contrastive learning which makes the learned structure similar to the original knowledge and the neglect of the expert knowledge model respectively. The simulation results are in agreement with our theoretical analysis. The model performs best when $r_{l}=0.8$ and $r_{a}=0.4$ in point of four metrics.

\subsection{Ablation Study}

\begin{figure}[htbp]
\centering
\includegraphics[width=0.47\textwidth]{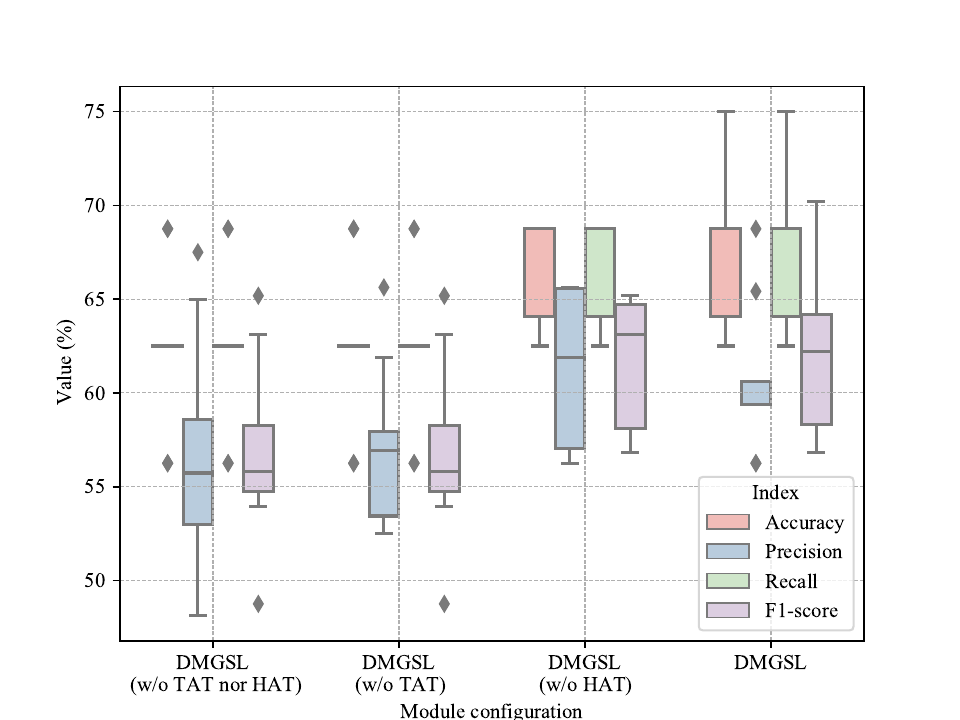}
\centering
\caption{Performance with different model configurations.}
\label{fig7}
\end{figure}

In order to verify the validity of the hierarchical attention model and temporal attention model introduced in our proposed method, we conduct an ablation study. Node classification performance is employed to evaluate if each component positively contributes to the final learned structure, as shown in Table \ref{Table3} and Fig. \ref{fig7}. First of all, considering only the hierarchical attention model (DMGSL (w/o TAT)), the classification accuracy is higher than other baselines and comparable to Sublime, but the comprehensive performance considering precision and F1-score is better than Sublime. Then, considering only the temporal attention model (DMGSL (w/o HAT)), the performance is significantly higher than the other baselines in the case of each indicator. Finally, the hierarchical attention model and temporal attention model are introduced together (DMGSL), it is obvious that the classification performance is further improved. Meanwhile, it can be seen in Fig. \ref{fig7} that when there is no attention models, the performance is relatively worst compared to other configurations. Therefore, both the hierarchical attention model and temporal attention model help enhance classification performance.

So far, we have demonstrated that DMGSL performs well on node classification. But whether it means good performance on structure learning? It is true to some degree. Although the process of node classification seems that only node attributes are paid attention, it is important to note that the impact of edges is also considered in the learning of the network structure and node embedding. Therefore, to a certain extent, the high performance achieved in the node classification task indicates that the learned network structure is meaningful. It not only incorporates the attributes of wireless network nodes but also captures the relations between nodes through edges, indeed supplementing and optimizing expert knowledge. By successfully constructing the network structure, our method provides a potential foundation for network automation.

\section{Conclusion}

The avenue of AI is fast evolving and will impact the telecom industry in the years to come. This study selects network automation as a cutting-in point and considers the problem of automatic graph representation for the next-generation mobile networks. Based on edge properties, the presented framework incorporates coherence time to partition dynamic networks into static snapshots. Hierarchical attention independently learns and merges slices, facilitates nuanced understanding of  relationships, while the temporal attention model integrates temporal features for enhanced learning. Furthermore, the method employs LSTM and multi-head attention mechanisms to capture temporal dynamics. Data augmentation techniques such as edge dropping and feature masking are utilized to mitigate overfitting and extract richer information from data. Overall, this research marks significant progress in autonomous learning and refinement of network topology for wireless communications, which paves the way for advancements in network automation.

\appendix

\section{A. Setup of experiments}
\label{setup}

The pre-processed network data, i.e., feature matrix, is input into the proposed framework together with the adjacency matrix of expert knowledge for training, which yields the output of the learned adjacency matrix and node embedding. To intuitively reflect the reliability of the learned structure, we use the learned node embedding for the node classification task. The training, validation, and test sets are divided by hierarchical sampling at a ratio of $6:2:2$.  To comprehensively evaluate the performance of node classification, we calculate four commonly used metrics: accuracy, precision, recall, and F1-score. The proposed method was implemented using an NVIDIA GeForce RTX 3060 Laptop GPU with $13.8\;{\rm{GB}}$ of memory. The implementation was based on PyTorch 1.9.0, utilizing the SGD and Adam optimizers.

\section{B. Dataset}
\label{dataset}

The final dataset can be summarized as follows, with details shown in Table~\ref{Table2} and Table~\ref{Table1}.
\begin{itemize}
	\item The adjacency matrix of the uplink throughput knowledge graph, generated by expert knowledge, includes $82$ nodes and $133$ relations, with three types of relations: causal relation (1), implicit relation (2), and explicit relation (3).
	\item The measured uplink throughput data (with $82$ data fields, including throughput capacity, block error rate, frame structure value, etc.) are collected with a window of 15 minutes and 35 minutes, resulting in $38,250$ and $120,418$ pieces of data, respectively. The data dimensions are $38,250 \times 82$ and $120,418 \times 82$.
\end{itemize}

\begin{table}[H]
	\begin{center}
		\begin{small}
			\begin{tabular}{cccc}
				\toprule
				\textbf{Data} & \textbf{Nodes} & \textbf{Edges} & \textbf{Features} \\
				\midrule
				\multirow{2}{*}{Adjacency matrix}  & $82$ & $133$ & \multirow{2}{*}{/} \\
				& ($10$ classes) & ($3$ classes) &  \\
				Uplink throughput & \multirow{2}{*}{/} & \multirow{2}{*}{/} & \multirow{2}{*}{$38250$} \\
				data (15min) & & & \\
				Uplink throughput & \multirow{2}{*}{/} & \multirow{2}{*}{/} & \multirow{2}{*}{$120418$} \\
				data (35min) & & & \\
				\bottomrule
			\end{tabular}
		\end{small}
	\end{center}
	\caption{Details of the wireless communication dataset.}
	\label{Table2}
\end{table}

\begin{table}[htbp]
	\renewcommand\arraystretch{1.2}
	\begin{center}
		\begin{small}
			\begin{tabular}{p{0.14\textwidth}p{0.115\textwidth}p{0.152\textwidth}}
				\toprule
				\textbf{Data field} & \textbf{Protocol layers} & \textbf{Data categories} \\
				\midrule
				nr$\_$pdsch$\_$bler             & physical layer  & block error rate      \\
				prb$\_$num$\_$ul$\_$slot        & physical layer  & frame structure value \\
				nr$\_$phy$\_$throughput$\_$ul   & data link layer & flow rate             \\
				nr$\_$rlc$\_$throughput$\_$ul   & data link layer & flow rate             \\
				nr$\_$ul$\_$dl$\_$slot$\_$ratio & network layer   & frame structure value \\
				nr$\_$dl$\_$slots$\_$pattern1   & network layer   & frame structure value \\
				\bottomrule
			\end{tabular}
		\end{small}
	\end{center}
	\caption{Examples of wireless communication data.}
	\label{Table1}
\end{table}

We perform data completion and normalization on the collected network data. The processed data are used as the feature matrix in structure learning.

\section{C. Influence of key hyperparameters}
\label{key hyperparameters}

\begin{figure}[!htbp]
	\centering
	\subfigure[Loss vs. Epoch with various $\tau$.]{
		\includegraphics[width=0.41\textwidth]{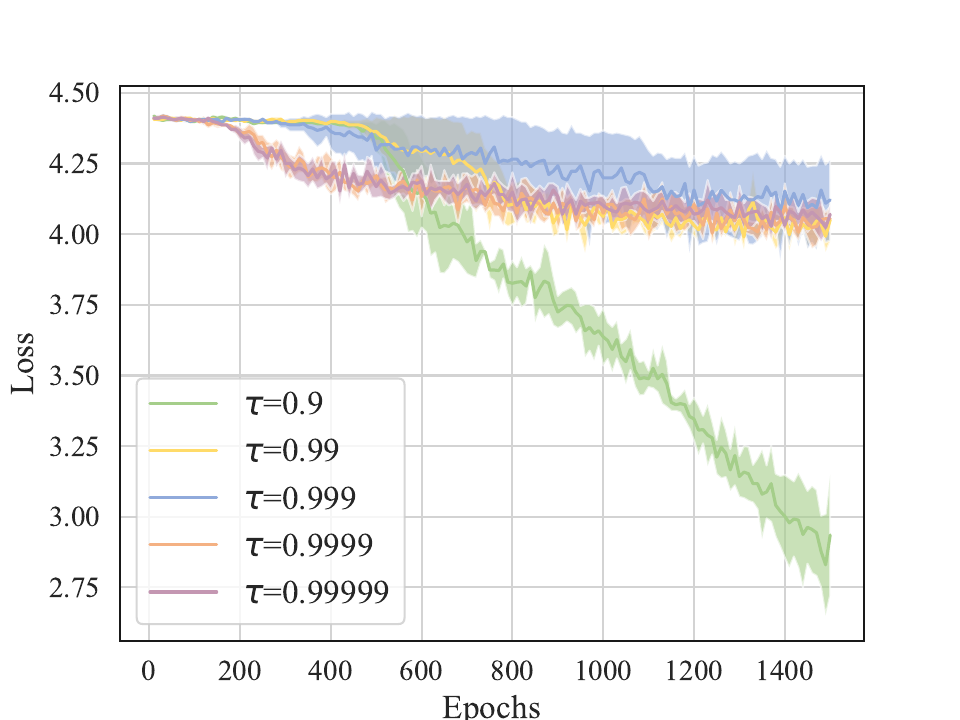}
		\label{fig4a}} 
	\subfigure[Loss vs. Epoch with various lr.]{
		\includegraphics[width=0.41\textwidth]{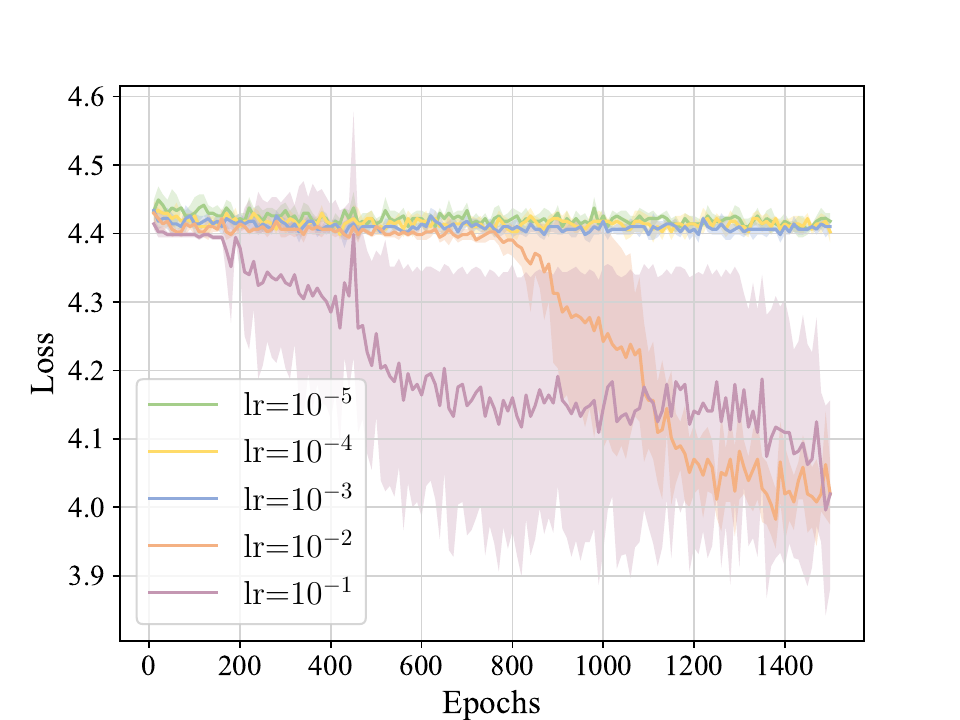}
		\label{fig4b}}
	\centering
	\caption{Training process.}
	\label{fig4}
\end{figure}

To prevent the learned graph from over-fitting the anchor graph (i.e., the learned adjacency matrix being too similar to the adjacency matrix of expert knowledge, thereby failing to learn new relations between data fields in the wireless network), a bootstrap mechanism is introduced in this method. This mechanism updates the anchor graph every $10$ epoches. $\tau\in[0,1]$ is a key parameter for adjusting the degree of updating: the closer $\tau$ is to $1$, the larger the proportion of the anchor graph in update process, meaning the degree of updating is lower. We varied the size of $\tau$ and conducted several experiments, plotting the curves of contrastive loss versus training epochs with different $\tau$ values, as shown in Fig.~\ref{fig4a}. We readily observe that, when $\tau$ is $0.9$, the contrastive loss shows a rapid downward trend initially but continues to decrease without a convergence trend as epochs increase. This is due to the rapid variation of the anchor graph, which causes unstable learning. When $\tau$ is greater than or equal to $0.99$, the curves gradually converge as epochs increase. The curve for $\tau=0.999$ converges the slowest; despite the curve for $\tau=0.99$ converges slowly at the beginning, it achieves a stable value faster than $\tau=0.9999$ and $\tau=0.99999$. Therefore, we conclude that when $\tau=0.99$, the model converges to a fair performance and is scalable.

\begin{figure}[!htbp]
	\centering
	\subfigure[Classification performance with different $\rm{lr}$.]{
		\includegraphics[width=0.41\textwidth]{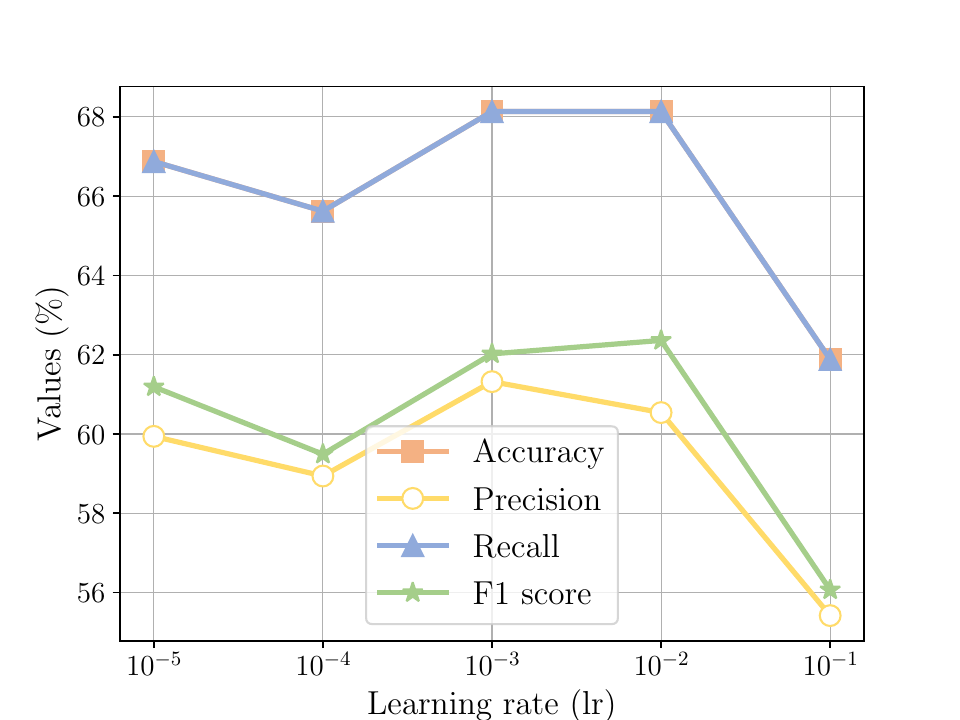}
		\label{fig5a}}
	\subfigure[Classification performance with different $k$.]{
		\includegraphics[width=0.41\textwidth]{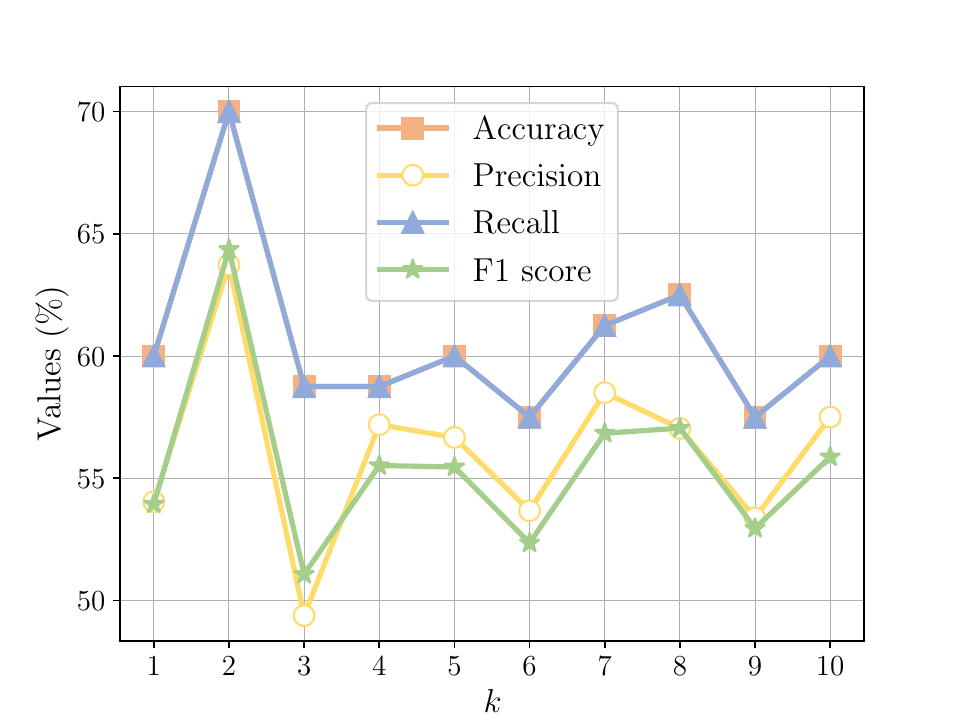}
		\label{fig5b}}
	\centering
	\caption{Influence of hyperparameters to classification performance.}
	\label{fig5}
\end{figure}

The learning rate, denoted as $\rm{lr}$, is another crucial parameter in deep learning, as it affects both the convergence speed and the effectiveness of learning. In order to find the optimal $\rm{lr}$, we set $\rm{lr}$ to [$10^{-1}$, $10^{-2}$, $10^{-3}$, $10^{-4}$, $10^{-5}$], and evaluated the classification performance. The curves of contrastive loss and classification performance with different $\rm{lr}$ values are shown in Fig.~\ref{fig4b} and Fig.~\ref{fig5a}, respectively. From \ref{fig4b} we see that, the curves for $\rm{lr} =10^{-5}$, $\rm{lr} =10^{-4}$, $\rm{lr} =10^{-3}$ show only minimal convergence trends as epochs increase. This is attributed to the convergence process slowed down by the small learning rate, which makes it difficult for the model to reach an optimal solution within a reasonable number of iterations. When $\rm{lr} =10^{-2}$, the curve stays stable at first and then decreases sharply until convergence, reaching a steady state faster than the curve for $\rm{lr} =10^{-1}$. From Fig. \ref{fig5a}, it is evident that $\rm{lr} =10^{-3}$ and $\rm{lr} =10^{-2}$ demonstrate comparable performance on node classification in terms of all four metrics. However, since the model with $\rm{lr} =10^{-3}$ does not converge, as analyzed previously, its performance on unseen data or in different scenarios cannot be guaranteed. Therefore, we choose $\rm{lr} =10^{-2}$ for rapid convergence.

When learning the structure of a mobile network, the nearest $k$ data fields of one data field are considered for topology learning. The value of $k$ influences the number of relationships that are ultimately learned. If $k$ is too small, the model may learn too few relationships and overlook important associations, such as indirect relations in the mobile network. Conversely, if $k$ is too large, it may acquire an excessive number of unnecessary relationships, leading to increased computational and memory demands. We set $k$ from $1$ to $10$ and conducted tests to obtain multiple groups of node classification performance. The curves of accuracy, precision, recall, and F1-score with different $k$ values are shown in Fig.~\ref{fig5b}. The results indicate that when $k$ is $2$, the performance of node classification is optimal. This suggests that with $k=2$, the adjacency matrix learned is the most reasonable in terms of node classification accuracy.

\bibliography{aaai24}

\end{document}